%% file: dahlia_icassp2027.tex
\documentclass{article}
\usepackage{spconf}

\usepackage{amsmath,amssymb,bm,mathtools}
\usepackage{graphicx}
\usepackage{xcolor}
\usepackage{booktabs}
\usepackage{colortbl,array}
\usepackage{multirow}
\usepackage{float}
\usepackage{placeins}
\usepackage{algorithm}
\usepackage{algpseudocode}
\usepackage[most]{tcolorbox}
\usepackage{listings}
\usepackage{comment}
\usepackage{url}
\usepackage{hyperref}

\hypersetup{
  pdftitle={Dahlia: An MLLM-Based Metric for Fine-Grained Text-Audio Alignment
            in Text-to-Audio Generation},
  pdfauthor={Shuitsu Koyama, Hiroki Nishimaki, Koki Seno, Yusuke Takagi,
             Hina Kogure, Yuiga Wada, Shunya Nagashima},
  colorlinks=false,
  pdfborder={0 0 0}
}

\newcommand{\para}[1]{%
  \vspace{1.2mm}
  \noindent\textbf{#1. }%
}

\newcommand{\TColorBox}[2]{%
\begin{tcolorbox}[enhanced,
  colback=blue!5!white,
  colframe=blue!75!black,
  fonttitle=\bfseries,
  title=#1,
  boxrule=0.5pt,
  sharp corners,
  left=2mm, right=2mm, top=1mm, bottom=1mm,
  before upper={\let\\\par},
]
#2
\end{tcolorbox}
}

\newcommand{\stdv}[2]{#1$\pm$#2}
\newcommand{\best}[1]{\textbf{#1}}
\newcommand{\second}[1]{\underline{#1}}

\title{Dahlia: An MLLM-Based Metric for Fine-Grained Text--Audio Alignment
       in Text-to-Audio Generation}

\name{\begin{tabular}{@{}c@{}}
  Shuitsu Koyama$^{1,2}$, Hiroki Nishimaki$^{2}$, Koki Seno$^{2}$,
  Yusuke Takagi$^{2}$, \\
  Hina Kogure$^{2}$, Yuiga Wada$^{2}$, Shunya Nagashima$^{1}$
\end{tabular}}
\address{$^{1}$Neurogica Inc. \qquad $^{2}$Keio University}

\begin{document}
\ninept
\maketitle

\input{report/abstract}


\input{report/introuction}
\input{report/related_work}
\input{report/problem_setting}
\input{report/experiment_setting}
\input{report/result}
\input{report/conclusion}

\clearpage

\bibliographystyle{IEEEbib}
\bibliography{mybib}

\end{document}

%% file: report/abstract.tex
\begin{abstract}
Text-to-audio generation has progressed rapidly toward realistic and diverse audio synthesis, yet reliable evaluation remains an open challenge. Widely used embedding-similarity metrics (e.g., CLAPScore) provide a useful signal for overall relevance but are not well-suited to fine-grained semantic alignment or compositional reasoning. To address this, we propose Dahlia, a multimodal large language model (MLLM)-based automatic evaluation metric that calibrates MLLM score distributions using embedding-based audio--caption alignment features. Specifically, Dahlia combines an Audio-to-Caption Alignment (A2C-Align) module with score distribution encoding and blending, leveraging both MLLM-based reasoning and similarity-based cross-modal evidence. Across XACLE, RELATE, and Clotho, Dahlia achieves the best SRCC among the compared metrics on all three benchmarks.
\end{abstract}

%% file: report/introuction.tex
\section{Introduction}
\vspace{-2mm}

Text-to-Audio (TTA) generation has been extensively studied and utilized across diverse real-world applications, including video game sound design~\cite{lloyd2011sound}, media production, and music generation.
To further advance the development of TTA models, it is essential to utilize reliable and scalable methods for assessing audio-text alignments (e.g., how well audio matches a prompt like ``keyboard typing sounds behind human speech.'')
Although human evaluation serves as the gold standard for reliability, it is labor-intensive and difficult to scale. Therefore, establishing automatic evaluation metrics that closely align with human judgments is crucial.

Most existing automatic metrics (e.g., CLAPScore \cite{xiao2024clapscore}) rely on a late-fusion design: they encode audio and text independently and compute a final score from the similarity of these global embeddings. Despite their scalability, such metrics often correlate poorly with human judgments.
This is partially because global embedding similarity cannot adequately capture fine-grained alignments between audio and texts, which results in overlooking intricate details~\cite{clap-not-fine1, elizalde2023clap, kanamori2025relate}.

Alternatively, early-fusion metrics (e.g., RELATE-Score \cite{kanamori2025relate}) perform cross-modal interactions at an earlier stage of the system.
For instance, RELATE-Score obtains embeddings from BYOL-A~\cite{byola} and RoBERTa~\cite{roberta} and then utilizes BLSTM to fuse them at an early stage.
While superior to late-fusion metrics on specific benchmarks, their performance can vary substantially across datasets.
This poor generalizability likely stems from the fact that both interaction modules (e.g., BLSTM) and embeddings (e.g., BYOL-A) are trained on highly specific distributions~\cite{polos}.
In light of these issues, automatic evaluation based on stronger and more generalizable models, such as multimodal large language models (MLLMs), would be a promising alternative.

Despite the potential of MLLMs, MLLM-based evaluation has not yet been fully investigated in this field. Recent MLLM-based evaluators provide stronger cross-modal reasoning than conventional embedding-similarity metrics. However, directly deriving scores from MLLM outputs remains unreliable: naively prompting the model to generate a numerical score correlates poorly with human judgments~\cite{geval, chan2023clair, lee-etal-2024-fleur, tong2025g-veval}, and score-token distributions can exhibit sample-dependent shape variation and numerical-token bias~\cite{symbolic}. Meanwhile, embedding-based representations, although insufficient on their own, provide a complementary signal of explicit audio–caption correspondence.

\begin{figure}[t]
    \centering
    \scalebox{1}{
        \includegraphics[width=\linewidth]{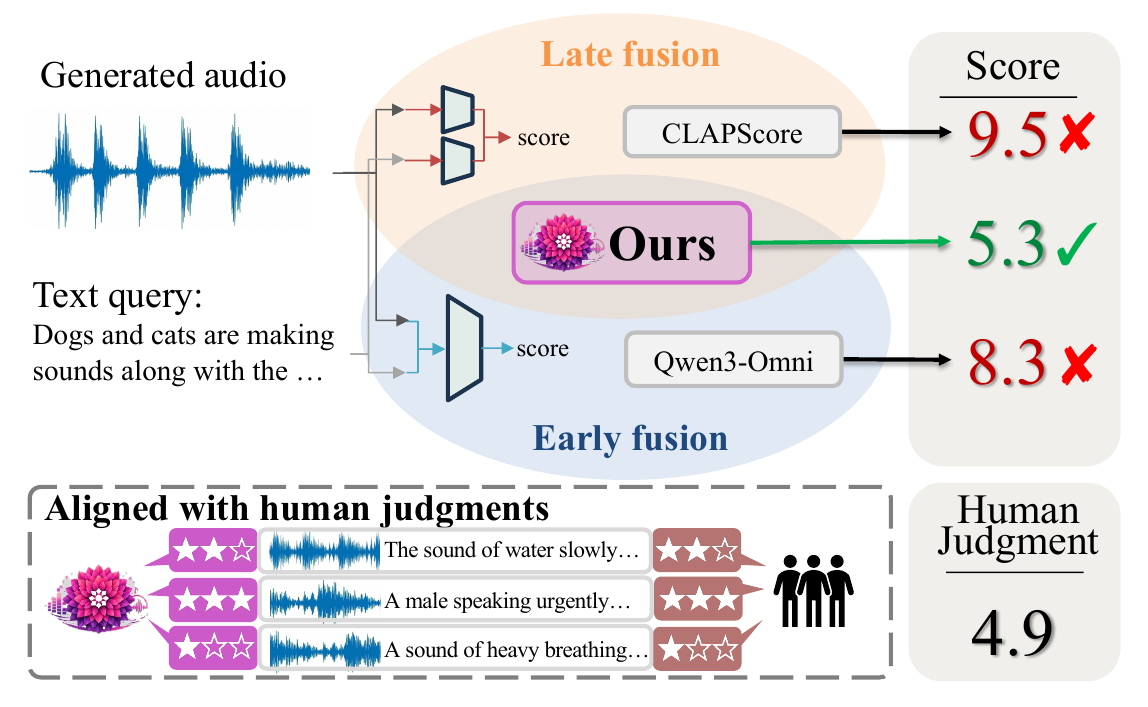}
    }
    \vspace{-5mm}
    \caption{Overview of Dahlia, an
MLLM-based automatic evaluation metric for TTA generation.
    Dahlia adopts a score-blending framework that aligns closely with human judgments by leveraging token-level logits.}
    \vspace{-7mm}
    \label{fig:eyecatch}
\end{figure}
\begin{figure*}[t]
    \centering
    \includegraphics[width=0.9\linewidth]{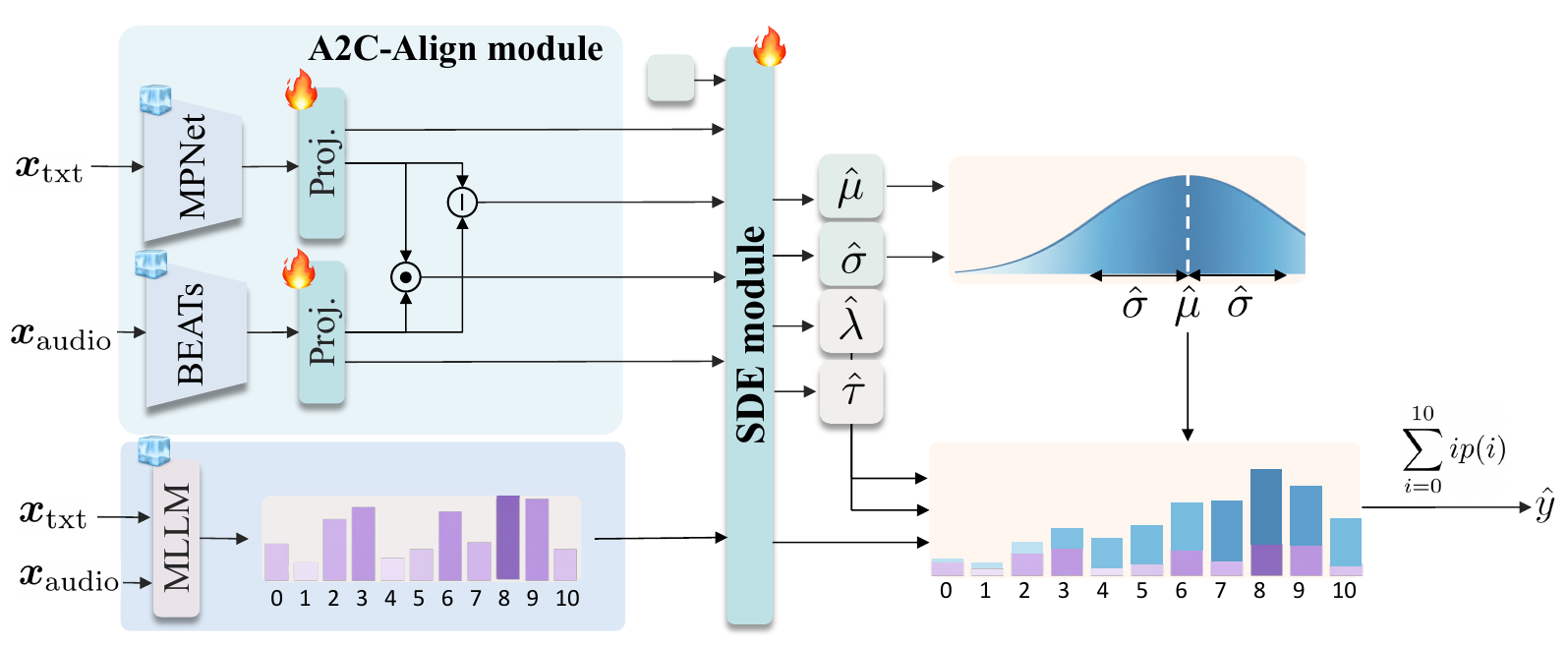}
    \vspace{-3mm}
    \caption{ Architecture of Dahlia. }
    \vspace{-6mm}
    \label{fig:model}
\end{figure*}
To address this, we propose Dahlia, an MLLM-based automatic evaluation metric that calibrates MLLM score distributions using explicit audio–caption alignment features. Dahlia employs an Audio-to-Caption Alignment (A2C-Align) module to extract embedding-based cross-modal evidence and a Score Distribution Encoding (SDE) module to jointly model this evidence with the MLLM score distribution. The resulting sample-specific parameters are used to calibrate the MLLM distribution through score blending before computing the final score. Fig.~\ref{fig:eyecatch} illustrates an overview of the proposed metric.

The main contributions are summarized as follows:
\begin{itemize}
    \item We propose Dahlia, an MLLM-based automatic evaluation metric for TTA generation that calibrates MLLM score distributions using explicit audio–caption alignment features.
    \item We introduce the A2C-Align module, the SDE module, and adaptive score blending, which jointly encode embedding-based alignment evidence with the MLLM score distribution and calibrate it in a sample-specific manner.
    \item Dahlia achieves the best SRCC among the compared metrics on XACLE~\cite{xacle2026challenge}, Clotho~\cite{kishi2025audiobertscore}, and RELATE~\cite{kanamori2025relate}.
\end{itemize}

%% file: report/related_work.tex
\vspace{-2mm}
\section{Related Work}
\vspace{-2mm}

\para{Automatic evaluation for text-to-audio generation}
Conventional text--audio evaluation metrics mainly rely on embedding similarity, such as CLAPScore~\cite{xiao2024clapscore, elizalde2023clap, wu2023laionclap}, or on learned cross-modal interaction modules, such as RELATE~\cite{kanamori2025relate}.
Human-CLAP~\cite{takano2025humanclap} fine-tunes CLAP on human judgments to improve correlation with subjective ratings.
Training-free metrics have also been explored: AudioBERTScore~\cite{kishi2025audiobertscore} scores a candidate against a reference from the similarity of audio embedding sequences, and ELSA~\cite{suzuki2026elsa} decomposes the generated audio into the acoustic events derived from the text query and measures event-level alignment without training or references.
Recently, MLLM-as-a-Judge approaches have emerged as a promising alternative by leveraging the reasoning capabilities of large multimodal models for automatic evaluation~\cite{geval, kuan2026aqascore, lee-etal-2024-fleur, tong2025g-veval, koyama2026rigel}.
For example, AQAScore~\cite{kuan2026aqascore} reformulates text--audio alignment evaluation as audio question answering and derives scores from the probability of a ``Yes'' response to semantic queries.
Building on this direction, Dahlia develops an MLLM-based evaluator for text--audio alignment that calibrates score-token logits through a learned score-blending framework.

%% file: report/problem_setting.tex
\vspace{-2mm}
\section{Problem Setting}
\vspace{-2mm}

We focus on the automatic evaluation of text-audio alignments of TTA models. Specifically, our focus is assessing models in terms of text–audio relevance (REL) (i.e., semantic and event-level alignments). We assume that our focus is REL rather than \textit{overall audio quality} \cite{kanamori2025relate} that targets perceptual attributes (e.g. {naturalness, fidelity}, and {recording quality}), as the latter do not directly consider how well generated audio aligns with prompts given to TTA models.
Furthermore, given that reference audio is not always available in real-world scenarios, reference-free evaluation is more practical than reference-based (e.g., \cite{kishi2025audiobertscore}); therefore, we adopt a reference-free setting.

Given a text prompt $x_{\mathrm{txt}}$ and a candidate audio $x_{\mathrm{audio}}$, automatic evaluation metrics assess alignments between $x_{\mathrm{txt}}$ and $x_{\mathrm{audio}}$ by outputting scores $\hat{y}$.
To validate metrics, we measure the correlation between automatic evaluation scores and human judgment scores.
Specifically, we employ Pearson linear correlation coefficient (LCC),  Spearman’s rank correlation coefficient (SRCC) and Kendall’s $\tau_b$, as they are standard in this field.

\vspace{-2mm}
\section{Method}
\vspace{-2mm}

We propose \textbf{Dahlia}, a reference-free MLLM-based automatic evaluation metric for Text-to-audio (TTA) models.
Fig. \ref{fig:model} shows the overview of the proposed metric.
Dahlia combines MLLM-based cross-modal reasoning with explicit audio–caption alignment features.

\input{tab/tables}
Dahlia consists of three main components: (i) an Audio-to-Caption Alignment (A2C-Align) module, which extracts explicit cross-modal alignment features from frozen audio and text encoders; (ii) a Score Distribution Encoding (SDE) module, which jointly models the MLLM score distribution and the alignment features along the discrete score axis; and (iii) score blending, which calibrates the MLLM score distribution using an adaptive prior and produces the final score by expectation.

Initially, we feed a prompt $x_{\text{prompt}}$ to the MLLM (Qwen3-Omni-30B-A3B-Instruct~\cite{xu2025qwen3omni}) to obtain the score-token logits $\hat{\bm{\ell}}_{\mathrm{MLLM}}\in\mathbb{R}^{K+1}$, where $K$ denotes the maximum score value. We set $K=10$ throughout all experiments. We write $p_{\mathrm{MLLM}}=\mathrm{softmax}(\hat{\bm{\ell}}_{\mathrm{MLLM}})$ for the corresponding score distribution, obtained by normalizing the logits over the $K+1$ score tokens.
This prompt is designed to enable the MLLM to evaluate text-audio relevance, focusing on the aspects emphasized in RELATE \cite{kanamori2025relate}.
Specifically, the prompt instructs the MLLM to assess (i) whether the audio contains the sound events described in the caption, and (ii) whether their temporal order is consistent with the caption when an order is specified.
Additionally, the base instruction of $x_{\text{prompt}}$ is inspired by previous works on LLM-as-a-judge~\cite{chan2023clair,lee-etal-2024-fleur,tong2025g-veval} that have demonstrated high evaluation performance in various domains.

\vspace{-2mm}
\subsection{Audio-to-Caption Alignment (A2C-Align) Module}
\vspace{-1mm}
To complement the MLLM score distribution with an explicit audio--caption alignment signal, we introduce the Audio-to-Caption Alignment (A2C-Align) module. Given candidate audio $\bm{x}_{\mathrm{audio}}$ and caption $\bm{x}_{\mathrm{txt}}$, we extract audio and text representations using frozen BEATs~\cite{pmlr-v202-chen23ag} and all-mpnet-base-v2~\cite{song2020mpnet}, respectively. The frame-level audio representations are mean-pooled, while the text representations are obtained by attention-mask mean pooling followed by L2 normalization.

Let $\bm{a}$ and $\bm{t}$ denote the pooled audio and text representations, respectively. We project them into a shared $d$-dimensional space as $\bm{c}=\mathrm{LN}(W_a\bm{a}+\bm{b}_a)$ and $\bm{r}=\mathrm{LN}(W_t\bm{t}+\bm{b}_t)$, where $\mathrm{LN}(\cdot)$ denotes layer normalization and $d=64$. We then construct four alignment tokens:
\vspace{-1mm}
\begin{equation}
\bm{A}_{\mathrm{A2C}}=[\,\bm{c};\,\bm{r};\,|\bm{c}-\bm{r}|;\,\bm{c}\odot\bm{r}\,]
\in\mathbb{R}^{4\times d}.
\end{equation}

\vspace{-2mm}
\subsection{Score Distribution Encoding (SDE) Module}
\vspace{-1mm}
To model the structure of the MLLM score distribution, we introduce the Score Distribution Encoding (SDE) module.
In LLM-as-a-Judge methods for other tasks (e.g., FLEUR~\cite{lee-etal-2024-fleur} and G-Eval~\cite{geval}), scores are commonly derived as the expectation over score-token logits.
However, the logit distribution can vary substantially across samples, including unimodal and multimodal shapes, and LLM-as-a-Judge systems may additionally exhibit symbolic bias, where logits for particular number tokens are spuriously amplified~\cite{symbolic}.
These effects can bias the expectation computed from the raw score distribution.

Rather than collapsing the score-token distribution into a scalar, SDE represents each score value as a token and jointly contextualizes the score distribution with the audio--caption alignment features from A2C-Align.
For each score $i=0,\ldots,K$, we construct a score token as
\vspace{-1mm}
\begin{equation}
\bm{s}_i=W_s\left[p_{\mathrm{MLLM}}(i);\,\hat{\ell}_{\mathrm{MLLM}}(i)/K;\,\bm{e}_i\right]+\bm{b}_s+\bm{u}_i,
\end{equation}
\vspace{-1mm}
where $\bm{e}_i$ denotes the one-hot representation of score $i$, $\bm{u}_i$ is a learned position embedding, and the logit is divided by $K$ for feature scaling.
A learned type embedding is additionally added to distinguish the {[}CLS{]}, A2C-Align, and score tokens.

The {[}CLS{]} token, four A2C-Align tokens, and $K+1$ score tokens are processed by a two-layer Transformer encoder.
From the resulting {[}CLS{]} representation, a linear layer predicts four raw parameters $\lambda_{\mathrm{raw}}$, $\mu_{\mathrm{raw}}$, $\sigma_{\mathrm{raw}}$, and $\tau_{\mathrm{raw}}$.
We obtain $\hat{\lambda}=1/(1+\mathrm{softplus}(\lambda_{\mathrm{raw}}))$, $\hat{\mu}=K\,\mathrm{sigmoid}(\mu_{\mathrm{raw}})$, $\hat{\sigma}=\mathrm{softplus}(\sigma_{\mathrm{raw}})$, and $\hat{\tau}=\mathrm{softplus}(\tau_{\mathrm{raw}})+10^{-3}$.
Here, $\hat{\mu}$ and $\hat{\sigma}$ determine the center and spread of the adaptive score prior, while $\hat{\lambda}$ controls the blending strength and $\hat{\tau}$ calibrates the sharpness of the MLLM score distribution.

\vspace{-2mm}
\subsection{Score Blending}
\vspace{-1mm}
Using the parameters predicted by SDE, we calibrate the MLLM score distribution before computing the final score.
To mitigate distribution-shape variation and symbolic bias in the score-token logits~\cite{symbolic}, we construct a sample-specific Gaussian prior $\hat{\bm{q}}=\{\hat{q}(i)\}_{i=0}^{K}$ parameterized by $\hat{\mu}$ and $\hat{\sigma}$:
\vspace{-1mm}
\begin{equation}
\hat{q}(i)= \frac{ \exp\!(-\frac{(i-\hat{\mu})^2}{2\hat{\sigma}^2}) }{ \sum_{j=0}^{K}\exp{\!(-\frac{(j-\hat{\mu})^2}{2\hat{\sigma}^2})}}, \quad i=0,\ldots,K.
\end{equation}
\vspace{-1mm}
The MLLM score distribution is then temperature-calibrated by $\hat{\tau}$ and blended with the adaptive prior using $\hat{\lambda}$:
\vspace{-1mm}
\begin{equation}
\hat{\ell}(i)=\hat{\lambda}\,\frac{\hat{\ell}_{\mathrm{MLLM}}(i)}{\hat{\tau}}
+(1-\hat{\lambda})\,\log\hat{q}(i), \quad i=0,\ldots,K,
\end{equation}
\vspace{-1mm}
where $\hat{\tau}$ rescales the sharpness of the MLLM score distribution and $\hat{\lambda}\in(0,1)$ is the mixing coefficient between the MLLM score distribution and the prior.
The final score $\hat{y}$ is obtained as the expectation of the calibrated distribution $p=\mathrm{softmax}(\hat{\ell})$:
\vspace{-2mm}
\begin{equation}
\hat{y}=\sum_{i=0}^{K} i\,p(i).
\end{equation}
 The model is trained with the squared error between the human judgment $y$ and the predicted score $\hat{y}$.

%% file: tab/tables.tex
\begin{table*}[t]
\centering
\small
\setlength{\tabcolsep}{4pt}
\renewcommand{\arraystretch}{1.1}
\caption{Quantitative results on the XACLE, RELATE, and Clotho test sets. In the table, $\rho$, $r$, and $\tau_b$ correspond to SRCC, LCC, and Kendall’s $\tau_b$, respectively. \textbf{Bold} indicates the best result in each column; \underline{underline} indicates the second best. ``--'' indicates either non-executable code or unavailable data.
}
\begin{tabular}{l *{9}{r}}
\toprule
\multirow{2}{*}{Metrics[\%]}
& \multicolumn{3}{c}{XACLE}
& \multicolumn{3}{c}{RELATE}
& \multicolumn{3}{c}{Clotho} \\
\cmidrule(lr){2-4}\cmidrule(lr){5-7}\cmidrule(lr){8-10}
& $\rho$ & $r$ & $\tau_b$
& $\rho$ & $r$ & $\tau_b$
& $\rho$ & $r$ & $\tau_b$ \\
\midrule
RELATE-Score
& 33.5 & 34.2 & 22.9
& 35.1 & 35.5 & 24.3
& 20.1 & 19.3 & 13.5 \\
PAM
& 20.9 & 22.9 & 14.2
& 19.9 & 19.9 & 13.5
& 13.3 & 13.4 & 9.5 \\
CLAPScore
& 33.3 & 37.2 & 22.8
& 19.9 & 22.7 & 13.6
& 41.3 & 41.8 & 28.5 \\
CLAPScore-MS
& 34.0 & 41.2 & 23.3
& 42.5 & 45.3 & 29.0
& 37.4 & 39.2 & 26.0 \\
CLAPScore-LAION
& 39.6 & 47.2 & 27.2
& 41.6 & 43.9 & 28.8
& 41.9 & 43.1 & 28.2 \\
Human-CLAPScore
& 42.6 & 49.3 & 29.5
& 46.1 & 49.0 & 32.4
& \second{50.7} & \second{51.0} & 35.6 \\
Qwen3-Omni~\cite{xu2025qwen3omni}
& 43.4 & 46.6 & 35.1
& 39.2 & 41.9 & 31.8
& 48.7 & 46.5 & \second{39.4} \\
AQAScore~\cite{kuan2026aqascore}
& -- & -- & --
& \second{55.6} & \second{54.4} & \second{39.6}
& -- & -- & -- \\
MoEScore~\cite{sun2026moescore}
& \second{64.0} & \best{68.7} & \second{46.1}
& -- & -- & --
& -- & -- & -- \\
\rowcolor{green!10}
Dahlia
& \best{\stdv{65.1}{0.2}} & \second{\stdv{67.1}{0.8}} & \best{\stdv{47.1}{0.2}}
& \best{\stdv{60.0}{0.6}} & \best{\stdv{60.5}{0.6}} & \best{\stdv{42.7}{0.5}}
& \best{\stdv{63.5}{0.8}} & \best{\stdv{62.6}{1.0}} & \best{\stdv{46.0}{0.6}} \\
\bottomrule
\end{tabular}
\label{tab:quantitative}
\end{table*}

%% file: report/experiment_setting.tex
\vspace{-2mm}
\section{Experimental Setting}
\vspace{-2mm}
\input{tab/ablation}
\para{Datasets and benchmarks}
We used standard benchmarks, including   XACLE~\cite{xacle2026challenge}, Clotho~\cite{kishi2025audiobertscore}, and RELATE~\cite{kanamori2025relate}.
We used the training sets to train the model, the validation sets to tune the hyperparameters, and the test sets to assess metric performance; for RELATE, we evaluated on the test-test set~\cite{kanamori2025relate}.

\para{Baselines}
As representative baselines, we employed RELATE-Score~\cite{kanamori2025relate}, PAM~\cite{deshmukh2024pam}, CLAPScore~\cite{xiao2024clapscore}\footnote{music\_speech\_audioset\_epoch\_15\_esc\_89.98.pt}, and Human-CLAPScore~\cite{takano2025humanclap}.
To examine the sensitivity of CLAPScore to the underlying CLAP backbone, we additionally considered two variants instantiated with different pretrained CLAP checkpoints: CLAPScore-MS, which uses MS-CLAP~\cite{elizalde2023clap}\footnote{the official 2023 pretrained weights (version="2023")}, and CLAPScore-LAION, which uses LAION-CLAP~\cite{wu2023laionclap}.\footnote{630k-audioset-best.pt}
Furthermore, Human-CLAPScore is a CLAPScore variant fine-tuned on the RELATE dataset with human judgments.
We further compared against AQAScore~\cite{kuan2026aqascore}, which scores alignment from the log-probability that an audio-aware LLM answers ``Yes'' to semantic queries, and MoEScore~\cite{sun2026moescore}, a mixture-of-experts evaluator submitted to the XACLE Challenge 2026. Finally, as an MLLM-as-a-Judge baseline, we use Qwen3-Omni~\cite{xu2025qwen3omni} directly: the backbone MLLM is given the same prompt $x_{\text{prompt}}$ and its generated score is used as the metric.

\para{Implementation details}
We adopted Qwen3-Omni-30B-A3B-Instruct~\cite{xu2025qwen3omni} as the backbone MLLM. The SDE module consists of a two-layer Transformer encoder with $d_{\mathrm{model}}=64$, four attention heads, an FFN dimension of 128, GELU activations, dropout of 0.1, and post-layer normalization. The projection dimension in A2C-Align is also set to 64. All pretrained encoders are kept frozen during training. We train the remaining parameters using Adam ($\beta_1=0.9$, $\beta_2=0.999$) with a learning rate of $10^{-3}$, weight decay of $10^{-4}$, and a batch size of 256 for up to 500 epochs. We apply early stopping with a patience of 50 and reduce the learning rate by a factor of 0.5 after 16 epochs without validation improvement. Results are averaged over three random seeds.

%% file: tab/ablation.tex
\begin{table*}[t]
\centering
\small
\setlength{\tabcolsep}{2pt}
\renewcommand{\arraystretch}{1.1}
\caption{Results of ablation studies. Values are reported as mean $\pm$ standard deviation over multiple runs. \textbf{Bold} indicates the best result in each column; \underline{underline} indicates the second best. }
\begin{tabular}{l *{9}{r}}
\toprule
\multirow{2}{*}{Metrics[\%]}
& \multicolumn{3}{c}{XACLE}
& \multicolumn{3}{c}{RELATE}
& \multicolumn{3}{c}{Clotho} \\
\cmidrule(lr){2-4}\cmidrule(lr){5-7}\cmidrule(lr){8-10}
& $\rho$ & $r$ & $\tau_b$
& $\rho$ & $r$ & $\tau_b$
& $\rho$ & $r$ & $\tau_b$ \\
\midrule

\rowcolor{green!10}
(a) Dahlia (full)
& \best{\stdv{65.1}{0.2}} & \best{\stdv{67.1}{0.8}} & \best{\stdv{47.1}{0.2}}
& \best{\stdv{60.0}{0.6}} & \best{\stdv{60.5}{0.6}} & \best{\stdv{42.7}{0.5}}
& \best{\stdv{63.5}{0.8}} & \best{\stdv{62.6}{1.0}} & \best{\stdv{46.0}{0.6}} \\
\midrule
(b) \ w/o score blending (direct, mean pool.)
& \stdv{63.5}{1.4} & \stdv{65.6}{2.5} & \stdv{45.5}{1.3}
& \stdv{58.7}{1.3} & \second{\stdv{58.9}{1.4}} & \stdv{41.6}{1.0}
& \stdv{59.3}{2.6} & \stdv{56.8}{3.4} & \stdv{42.4}{2.2} \\
(c) \ w/o score blending (direct, CLS)
& \second{\stdv{63.7}{0.7}} & \second{\stdv{65.9}{1.8}} & \second{\stdv{45.6}{0.7}}
& \second{\stdv{59.0}{1.2}} & \second{\stdv{58.9}{1.5}} & \second{\stdv{41.9}{1.0}}
& \second{\stdv{60.4}{2.7}} & \second{\stdv{57.8}{3.9}} & \second{\stdv{43.2}{2.3}} \\
(d) \ w/o MLLM score distribution (uniform)
& \stdv{48.4}{0.3} & \stdv{49.0}{0.2} & \stdv{33.7}{0.3}
& \stdv{51.8}{0.5} & \stdv{49.7}{0.4} & \stdv{36.2}{0.3}
& \stdv{45.5}{2.9} & \stdv{43.8}{2.8} & \stdv{31.6}{2.2} \\
(e) \ w/o A2C-Align (logits only)
& \stdv{56.7}{0.1} & \stdv{58.6}{0.2} & \stdv{40.2}{0.1}
& \stdv{51.4}{0.3} & \stdv{52.3}{0.4} & \stdv{36.2}{0.3}
& \stdv{56.5}{0.6} & \stdv{55.5}{0.9} & \stdv{40.3}{0.5} \\
\bottomrule
\end{tabular}
\label{tab:ablation}
\end{table*}

%% file: report/result.tex
\vspace{-2mm}
\section{Results}
\para{Quantitative results}
Table \ref{tab:quantitative} shows the quantitative results.
On XACLE, Dahlia obtains an SRCC ($\rho$) of 65.1, an LCC ($r$) of 67.1, and a Kendall’s $\tau_b$ of 47.1. Compared with MoEScore~\cite{sun2026moescore}, the strongest baseline on this benchmark, Dahlia improves SRCC by 1.1 points and Kendall’s $\tau_b$ by 1.0 points, while its LCC is 1.6 points lower.
On RELATE, Dahlia reaches 60.0 in SRCC, 60.5 in LCC, and 42.7 in Kendall’s $\tau_b$, corresponding to gains of 4.4, 6.1, and 3.1 points over the strongest baseline, AQAScore~\cite{kuan2026aqascore}, respectively.
On Clotho, Dahlia records an SRCC of 63.5, an LCC of 62.6, and a Kendall’s $\tau_b$ of 46.0, exceeding the strongest baseline, Human-CLAPScore, by 12.8, 11.6, and 10.4 points, respectively.
Overall, Dahlia obtained the best score in eight of the nine columns and the best SRCC on all three benchmarks, indicating strong agreement with human judgments across diverse TTA evaluation benchmarks.

\para{Ablation studies}
Table~\ref{tab:ablation} presents the ablation results. Variant (a) is the full model; (b) and (c) drop score blending and regress the score directly from the fused representation via mean pooling and the CLS token, respectively; (d) removes the MLLM score evidence by replacing $\hat{\bm{\ell}}_{\mathrm{MLLM}}$ with a uniform distribution, leaving SDE to predict the prior parameters from the A2C-Align tokens alone; and (e) removes A2C-Align and uses the MLLM score distribution alone.

Both sources of evidence contributed substantially to performance: removing the MLLM score distribution (d) or A2C-Align (e) degraded SRCC by up to 18.0 and 8.6 points, respectively. Score blending also contributed: against the stronger direct-regression variant (c), the full model gained 1.4, 1.0, and 3.1 points in SRCC and 1.2, 1.6, and 4.8 points in LCC on XACLE, RELATE, and Clotho, respectively, and markedly reduced run-to-run variance (e.g., an LCC standard deviation of 1.0 on Clotho against 3.9 for (c)).

\begin{figure}[t]
    \centering
    \includegraphics[clip,width=\linewidth]{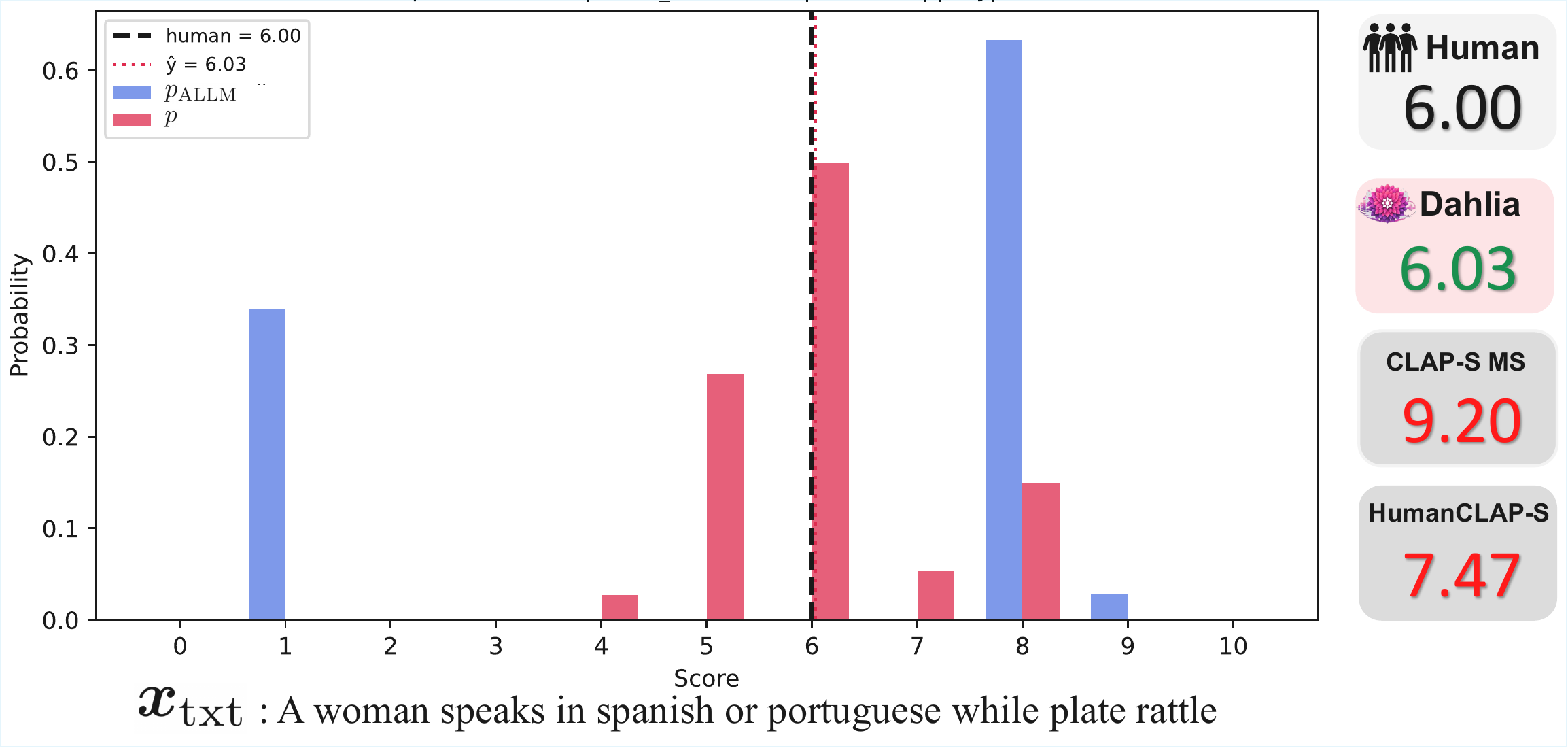}
    \vspace{-6mm}
    \caption{ Qualitative results on the XACLE dataset. }
    \label{fig:p_tildep}
    \vspace{-7mm}
\end{figure}

\para{Qualitative results}
Fig. \ref{fig:p_tildep} presents a representative success case of Dahlia on XACLE. In this example, the generated audio did not contain the target “plate rattle” sound; instead, it resembled “keyboard typing”. Moreover, the sound occurs after the speech segment, which is inconsistent with the phrase “while plate rattle” and therefore mismatches the caption at the level of event ordering. This might lead annotators to give a score of 6.00. CLAPScore-MS and Human-CLAPScore gave a score of 9.20 and 7.47, respectively, whereas Dahlia gave 6.03, closely matching the human judgment.

Furthermore, Fig. \ref{fig:p_tildep} compares the score-token distributions before and after calibration to analyze the source of this improvement. Let $p_{\mathrm{MLLM}}=\mathrm{softmax}(\hat{\ell}_{\mathrm{MLLM}})$ denote the raw distribution over score tokens. As shown in Fig. \ref{fig:p_tildep}, $p_{\mathrm{MLLM}}$ is multimodal, with disproportionately high mass on two extreme outcomes (e.g., scores 1 and 8). In contrast, $p$ becomes unimodal, yielding an expected value of 6.03. This result indicates that Dahlia corrected the distribution shape.

%% file: report/conclusion.tex
\vspace{-2mm}
\section{Conclusion}
\vspace{-2mm}
The main contributions are summarized as follows:
    (i) We proposed Dahlia, an MLLM-based automatic evaluation metric for TTA generation that calibrates MLLM score distributions using explicit audio–caption alignment features.
    (ii) Dahlia combines an A2C-Align module, an SDE module, and adaptive score blending to calibrate the MLLM score distribution in a sample-specific manner.
    (iii) Dahlia achieved the best SRCC among the compared metrics on XACLE, Clotho, and RELATE.